\documentclass[acmtog]{acmart}
\acmSubmissionID{1054} %

\usepackage{booktabs} %

\usepackage[ruled]{algorithm2e} %

\SetAlFnt{\small}
\SetAlCapFnt{\small}
\SetAlCapNameFnt{\small}
\SetAlCapHSkip{0pt}

\usepackage{tabularx} %
\usepackage{multirow} %
\usepackage{cleveref}

\makeatother

\newcolumntype{Y}{>{\centering\arraybackslash}X}
\newcolumntype{P}[1]{>{\centering\arraybackslash}p{#1}}
\newcolumntype{C}{>{\centering}X}

\newcommand{\OURS}{GGHead}

\acmJournal{TOG}

\copyrightyear{2024}
\acmYear{2024}
\setcopyright{rightsretained}
\acmConference[SA Conference Papers '24]{SIGGRAPH Asia 2024 Conference Papers}{December 3--6, 2024}{Tokyo, Japan}
\acmBooktitle{SIGGRAPH Asia 2024 Conference Papers (SA Conference Papers '24), December 3--6, 2024, Tokyo, Japan}\acmDOI{10.1145/3680528.3687686}
\acmISBN{979-8-4007-1131-2/24/12}

\begin{document}
\title{GGHead: Fast and Generalizable 3D Gaussian Heads}

\author{Tobias Kirschstein}
\affiliation{%
  \institution{Technical University of Munich}
  \country{Germany}
}
\author{Simon Giebenhain}
\affiliation{%
  \institution{Technical University of Munich}
  \country{Germany}
}
\author{Jiapeng Tang}
\affiliation{%
  \institution{Technical University of Munich}
  \country{Germany}
}
\author{Markos Georgopoulos}
\affiliation{%
  \institution{Independent Researcher}
  \country{Switzerland}
}
\author{Matthias Nießner}
\affiliation{%
  \institution{Technical University of Munich}
  \country{Germany}
}

\begin{abstract}
Learning 3D head priors from large 2D image collections is an important step towards high-quality 3D-aware human modeling.
A core requirement is an efficient architecture that scales well to large-scale datasets and large image resolutions.
Unfortunately, existing 3D GANs struggle to scale to generating samples at high resolutions due to their relatively slow train and render speeds, and typically have to rely on 2D superresolution networks at the expense of global 3D consistency.
To address these challenges, we propose Generative Gaussian Heads (GGHead), which adopts the recent 3D Gaussian Splatting representation within a 3D GAN framework.
To generate a 3D representation, we employ a powerful 2D CNN generator to predict Gaussian attributes in the UV space of a template head mesh.
This way, GGHead exploits the regularity of the template's UV layout, substantially facilitating the challenging task of predicting an unstructured set of 3D Gaussians. 
We further improve the geometric fidelity of the generated 3D representations with a novel total variation loss on rendered UV coordinates. 
Intuitively, this regularization encourages that neighboring rendered pixels should stem from neighboring Gaussians in the template's UV space.
Taken together, our pipeline can efficiently generate 3D heads trained only from single-view 2D image observations.
Our proposed framework matches the quality of existing 3D head GANs on FFHQ while being both substantially faster and fully 3D consistent.
As a result, we demonstrate real-time generation and rendering of high-quality 3D-consistent heads at $1024^2$ resolution for the first time.

Project Website: {\color[HTML]{0065bd}\href{https://tobias-kirschstein.github.io/gghead}{\texttt{https://tobias-kirschstein.github.io/gghead}}}

\end{abstract}

\begin{CCSXML}
<ccs2012>
    <concept>
    <concept_id>10010147.10010178.10010224.10010245.10010254</concept_id>
    <concept_desc>Computing methodologies~Reconstruction</concept_desc>
    <concept_significance>500</concept_significance>
</concept>
<concept>
    <concept_id>10010147.10010178.10010224.10010226.10010239</concept_id>
    <concept_desc>Computing methodologies~3D imaging</concept_desc>
    <concept_significance>500</concept_significance>
</concept>
    <concept>
    <concept_id>10010147.10010257.10010258.10010261.10010276</concept_id>
    <concept_desc>Computing methodologies~Adversarial learning</concept_desc>
    <concept_significance>500</concept_significance>
</concept>
</ccs2012>
\end{CCSXML}

\ccsdesc[500]{Computing methodologies~Reconstruction}
\ccsdesc[500]{Computing methodologies~3D imaging}
\ccsdesc[500]{Computing methodologies~Adversarial learning}

\keywords{3D GAN, 3D head prior, 3D Gaussian Splatting}

\begin{teaserfigure}
  \includegraphics[width=\textwidth]{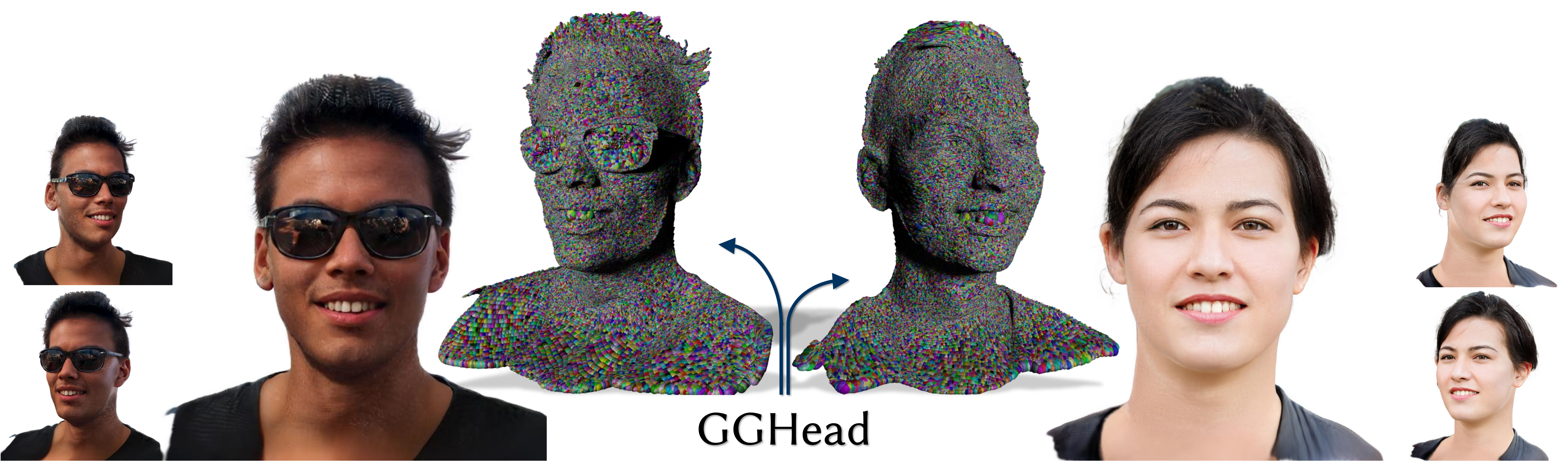}
  \caption{\textbf{GGHead:} Our method can generate diverse 3D head representations based on 3D Gaussian Splatting. The sampled persons exhibit detailed geometry and appearance. Both generation and rendering can be conducted in real-time at full image resolution without the need for 2D super-resolution networks.}
  \label{fig:teaser}
\end{teaserfigure}

\maketitle
\section{Introduction}

A high-quality 3D generative model that can sample human heads with diverse geometry and appearance has to satisfy two important constraints: A high rendering resolution and strict 3D consistency of generated imagery. Once met, these properties enable exciting use-cases such as single-image-to-3D reconstruction, 3D content creation, or 3D-consistent editing of images.

Unfortunately, obtaining such a high-quality 3D prior over human appearances is extremely challenging. Not only does such a model have to explain all diverse human appearances, but there is also a serious lack of large-scale 3D human datasets due to expensive data capture procedures, rendering any attempt to learn 3D human priors from 3D datasets alone impractical. Hence, methods need to be able to learn 3D appearance and geometry from large 2D image collections, requiring them to render during training. As a result, to meet the criteria for quality and 3D-consistency, the employed rendering procedure has to be scalable, both in terms of speed and rendering resolution.

A well-established approach are 3D Generative Adversarial Networks (3D-GANs) that can learn unconditional 3D generation models from 2D images with corresponding camera poses. These methods have shown remarkable 3D-aware image synthesis capabilities, rapidly closing the quality gap to methods that directly generate 2D face images without view control. Under the hood, 3D GANs combine a differentiable rendering pipeline with a view-conditioned discriminator architecture to supervise 3D generation solely with 2D images. 
To obtain high-quality 3D samples with this approach, a high rendering resolution is crucial. A commonly chosen and successful differentiable rendering pipeline is Neural Radiance Fields (NeRFs)~\cite{mildenhall2021nerf}. However, so far, scaling the rendering resolution of NeRFs has been bottle-necked by the expensive ray marching procedure necessary to synthesize the final image. Several works have attempted to alleviate this limitation: EG3D~\cite{chan2022eg3d} renders images at lower resolutions and then upsamples them using 2D superresolution, sacrificing 3D consistency in the process. GRAM~\cite{deng2022gram} and GRAM-HD~\cite{xiang2023gramhd} speed up ray marching by simplifying the 3D representation to a set of 2D manifolds, allowing cheap ray intersection tests. However, the simplistic 3D representation worsens rendering quality especially for side-views. Mimic3D~\cite{chen2023mimic3d} generates a pseudo-GT dataset from EG3D to supervise the 3D representation directly with patch-based rendering. Due to the inherent view-inconsistencies in the generated multi-view images from EG3D, only a perceptual loss can be used for supervision to gloss over the misalignment. Most recently, \citet{trevithick2024wysiwyg} investigate how ray marching can be sped up with proposal networks, allowing training at $512^2$ resolution. 
Nevertheless, rendering speeds during both training and inference remain a significant challenge.
For instance, currently, none of the existing methods are capable of natively generating and rendering images at full $512^2$ resolution in real-time. Since the creation of a 3D GAN involves several million render passes, this imposes a considerable computational burden on training.

To address these  scalability challenges, we propose to replace the commonly used NeRF 3D representation with the recent 3D Gaussian Splatting (3DGS) pipeline~\cite{kerbl2023gaussiansplatting}. Based on rasterization instead of expensive ray marching, 3DGS has shown remarkable rendering speeds harboring great potential for more scalable 3D GAN pipelines. To generate a set of 3D Gaussian primitives, we exploit the UV space of a template mesh, allowing us to leverage powerful 2D CNN architectures such as StyleGAN2~\cite{karras2020stylegan2}. From the predicted 2D maps for position, scale, rotation, color, and opacity, we sample 3D Gaussians and place them in the 3D scene relative to the template. Finally, the 3D representation is rendered and supervised with a view-point aware discriminator. 
The purpose of the template mesh is two-fold: (i) Simplify the generation process by predicting structured 2D UV maps instead of unstructured 3D primitives, (ii) Regularize the predicted 3D representation. The latter is necessary because 3D Gaussians, as opposed to radiance fields, are highly sensitive to just a few parameters. Therefore, bad gradients during the early stages of adversarial training can cause blow-up. To counteract this, we propose a novel UV total variation loss that makes the 3D representation more well-behaved by enforcing continuity of rendered pixels with respect to their UV coordinates. Our final model can learn 3D head geometry and appearance priors from 2D data collections alone. 
Furthermore, {\OURS} generates and renders photorealistic 3D heads in real-time thanks to its efficient 3D representation, allowing it to scale to larger resolutions.

In summary, our contributions are as follows:
\begin{itemize}
    \item We propose {\OURS}, a 3D GAN for human heads based on 3D Gaussian Splatting that is trained only from 2D image observations.
    \item We parameterize 3D Gaussian Heads as UV maps that can be generated with efficient 2D CNNs in conjunction with a novel UV total variation regularization that improves geometric fidelity.
    \item Our approach is highly scalable, facilitating training at 1k resolutions while enabling both real-time sample generation and head rendering.
\end{itemize}

\section{Related Work}
In this section, we discuss prior related work in the field of generative modeling of 3D heads. We begin by reviewing existing 3D GANs and their application to 3D heads. We then discuss 3DGS-based models and their extension to generalizable representations.

\subsection{3D Generative adversarial models}
Following the success of their 2D counterpart, 3D GANs employ adversarial training to obtain a generative model of 3D representations from 2D images.
Earlier works utilise implicit 3D representations and directly render pixels  \cite{chan2021pi,schwarz2020graf} or features that are decoded with a CNN-based neural renderer \cite{niemeyer2021giraffe,xue2022giraffe}.
The latter approach has also been adopted by efforts that aim to repurpose successful 2D GAN architectures (e.g., \cite{karras2019style,karras2020stylegan2,Karras2020ada}) and lift the representation to 3D using implicit neural representations \cite{chan2022eg3d,or2022stylesdf,gu2021stylenerf}. 
Due to their success, these approaches have been extended to introduce different modalities of control, e.g., semantics \cite{sun2022ide}.

To obtain higher resolutions, some methods such as EG3D~\cite{chan2022eg3d} apply a super-resolution network to the rendered features. While efficient, this approach can introduce unwanted artefacts in the form of 3D inconsistencies as well as low-resolution geometry.
To alleviate this issue, several works focus on rendering in the pixel-space while maintaining fidelity and efficiency.
Among them, \citet{skorokhodov2022epigraf} propose to render patches instead of full images to reduce the computational cost.
While this allows supervising a generative 3D representation without super-resolution, it impacts global consistency due to the discriminator's limited field of view. 
Mimic3D~\cite{chen2023mimic3d} proposes an imitation strategy that forces the 3D representations to mimic the result of the 2D super-resolution branch.
Similarly, GRAM-HD~\cite{xiang2023gramhd} utilizes radiance manifolds and performs super-resolution on 2D manifolds instead of the rendered images.
More recently, \citet{trevithick2024you} introduce a learning-based sampling strategy to effectively render every pixel and to model high-resolution geometry.
Orthogonally to these work, our method aims to improve rendering speed by adopting 3D Gaussian Splatting as a 3D representation instead of NeRFs.

\subsection{Generalizable 3D Gaussian Splatting}
To mitigate the computational overhead of volumetric rendering imposed by implicit 3D representations, \citet{kerbl2023gaussiansplatting} introduced the seminal work of 3DGS.
In this work, a set of Gaussians is optimized from multi-view images using volume splatting \cite{zwicker2001ewa}. 
The means of the Gaussians are initialized using points from SfM \cite{schonberger2016structure} while subsequent efforts try to break this dependency \cite{fu2023colmap}.
Inspired by its efficient rendering, a plethora of methods have been proposed to extend 3DGS for real-time human avatar modeling utilising 3DMMs \cite{qian20233dgs,qian2023gaussianavatars,xu2023gaussian,zielonka2023drivable,jiang2024uv,svitov2024haha,hu2023gauhuman,moreau2023human,saito2023relightable}.
Although these efforts focus on modeling a single scene, 3DGS has been extended to generalizable methods for scene reconstruction \cite{xu2024grm,zou2023triplane}, as well as text-driven synthesis using Score Distillation Sampling \cite{yi2023gaussiandreamer,tang2023dreamgaussian,li2023gaussiandiffusion,ren2023dreamgaussian4d,ling2023align}.
Another line of work investigates the use of 3DGS for generalizable few-shot 3D reconstruction~\cite{wewer2024latentsplat, charatan2023pixelsplat}.
However, none of these works focus on learning an unconditional 3D generative model (e.g., a GAN) directly on 3DGS-based representations. Closer to our work, Gaussian Shell Maps (GSM)~\cite{abdal2023gsm} introduces a Gaussian-based 3D GAN for human bodies by relying on shellmaps of fixed Gaussians.
In contrast, we propose a 3D GAN for human heads without the use of shell maps or an animatable mesh prior, and instead with freely moving Gaussians and a novel regularization scheme that exploits the UV space of a template mesh.

\section{Method}

\begin{figure*}[htb]
  \includegraphics[width=\linewidth]{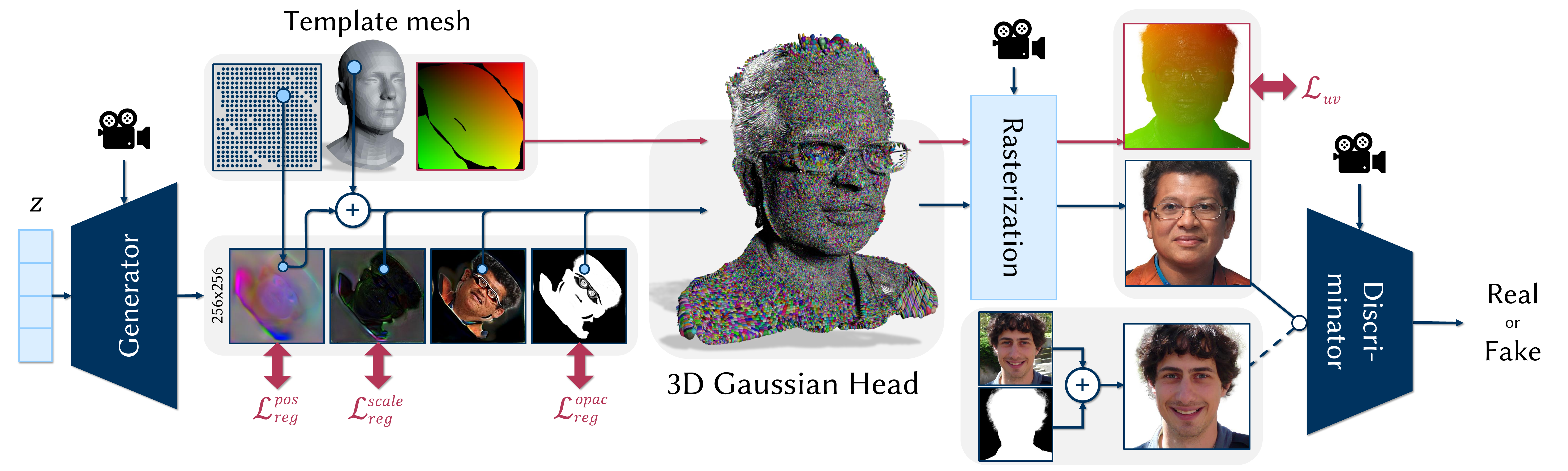}
  \caption{\textbf{Method Overview:} We adopt 3D Gaussian Splatting inside a 3D GAN formulation to learn a distribution of 3D heads solely from 2D images. To build the 3D Gaussian representation, we exploit the UV space of a template mesh. The CNN generator then takes a normally distributed latent code $z$ and predicts 2D maps for each Gaussian attribute. To obtain the actual 3D Gaussian primitives for rasterization, we sample the UV maps uniformly and place the primitives relative to the template mesh. The generated 3D Gaussian representation is then rasterized and supervised by the discriminator. To increase training stability, especially during early stages of adversarial training, we regularize the predicted position offsets, scales and opacities via $\mathcal{L}_{reg}^{pos}$, $\mathcal{L}_{reg}^{scale}$ and $\mathcal{L}_{reg}^{opac}$. Furthermore, we propose a novel UV total variation loss $\mathcal{L}_{uv}$ to improve the geometric fidelity of generated 3D heads by enforcing UV renderings to be smooth.
} 
  \label{fig:03_method_overview}
\end{figure*}

\Cref{fig:03_method_overview} shows an overview of our proposed pipeline. We pursue a 3D GAN approach, consisting of a generator that predicts 3D representations, a differentiable rasterizer, and a discriminator that supervises the image formation process. To make the setup scalable, we employ 3D Gaussian Splatting as an underlying 3D representation. In the following, we will discuss how we use a template mesh in conjunction with a 2D CNN to predict 3D Gaussian heads (\cref{sec:03_template_based_3d_gaussian_generation}), and how we regularize the 3D representation to ensure stable training in an adversarial setting (\cref{sec:03_gaussian_attribute_modeling,sec:03_gaussian_geometry_reg}).

\subsection{3D Gaussian Splatting (3DGS)}

3D Gaussian Splatting~\cite{kerbl2023gaussiansplatting} is a point-based scene representation that assigns each point five different attributes: The Gaussian center $\mu \in \mathbb{R}^3$, scale $\mathbf{s} \in \mathbb{R}^3$, rotation parameterized as quaternion $\mathbf{q} \in \mathbb{R}^4$, color $\mathbf{c} \in \mathbb{R}^3$, and opacity $\sigma \in \mathbb{R}$.
\begin{align}
    \mathcal{G}^i = \{\mathbf{\mu}, \mathbf{s}, \mathbf{q}, \mathbf{c}, \sigma\}
\end{align}

The resulting annotated point cloud can then be efficiently rendered into an image $I$ using the differentiable tile-based rasterizer $\mathcal{R}$ and camera parameters $\pi$:
\begin{align}
    I = \mathcal{R}(\mathcal{G}, \pi)
\end{align}
In the following, we will discuss how one can efficiently generate a 3D Gaussian pointcloud $\mathcal{G}$ using 2D CNN architectures.

\subsection{Template-based 3D Gaussian Generation}
\label{sec:03_template_based_3d_gaussian_generation}
3D Gaussians, being a point-based representation, are inherently unstructured. This poses a significant challenge for generation tasks due to issues such as order ambiguity or large regions of empty space. 
We therefore follow Gaussian Shell Maps~\cite{abdal2023gsm} and Relightable Gaussian Codec Avatars~\cite{saito2023relightable}, and rig the 3D Gaussian representation to a template mesh with corresponding UV layout. This enables the use of an efficient 2D CNN backbone $\mathcal{B}$ to predict the Gaussian representation $\mathcal{G}$, considerably simplifying the generator's task. Formally, we generate one UV map $M$ for each Gaussian attribute:
\begin{align}
    M_\star = \mathcal{B}(z, \pi)
\end{align}
where $\star \in \{\textsc{position},\textsc{scale}, \textsc{rotation}, \textsc{color}, \textsc{opacity}\}$ and $\mathcal{B}$ is a StyleGAN2~\cite{karras2020stylegan2} generator, mapping a normally distributed latent code $z\in \mathbb{R}^{512}$ to UV maps $M \in \mathbb{R}^{256 \times 256 \times 14}$.

For predicted positions, we add another learnable CNN layer $\mathcal{Z}$ with weights initialized as zeros following~\citet{zhang2023controlnet}: 
\begin{align}
    M_{position} \leftarrow \mathcal{Z}(M_{position})
\end{align}
This ensures that predicted position offsets are 0 at the beginning of training, causing Gaussians to be predicted on the template mesh. This is part of a set of measures that improve training stability as further elaborated in~\cref{sec:03_gaussian_geometry_reg,sec:03_gaussian_attribute_modeling}.

To obtain the actual 3D Gaussian primitives, we uniformly sample the predicted 2D maps $M$. Each Gaussian $\mathcal{G}^i$ has a fixed coordinate $x_{uv}^i \in [0, 1]^2$ in the template's UV space and will query its attributes from $M$:
\begin{align*}
    \mathcal{G} = \textsc{GridSample}(M, x_{uv})
\end{align*}
where the $\textsc{GridSample}(\cdot)$ operater performs lookup in the discrete maps $M$ via bilinear interpolation.
The sampling scheme governs how many Gaussians will be created and is shared across all generated persons. Note that the generator is mostly agnostic to how many Gaussians are sampled meaning that the sampling scheme may be changed during training, e.g., to increase the number of Gaussians when rendering at higher resolutions. 

\subsection{Gaussian Attribute Modeling}
\label{sec:03_gaussian_attribute_modeling}
\subsubsection{Gaussian Positions}
To ensure that the predicted Gaussian attributes always stay within a reasonable domain, we employ activation functions. Most notably, for the predicted positions, we add the Gaussian's corresponding 3D position $v^i$ on the template mesh and limit the predicted offset with a $\textsc{tanh}$ activation function:
\begin{align}
    \mathcal{G}_{position}^i &\leftarrow v^i + \gamma_{pos} \textsc{tanh}\left(\mathcal{G}_{position} ^i\right)
\end{align}
where $\gamma_{pos}$ is the maximum allowed predicted offset from the template mesh. For our experiments on FFHQ, we use $\gamma_{pos} = 0.25$ meaning that Gaussians can move at most 25cm away from the template.
This step is crucial, as it prevents the Gaussians from leaving the training view frustums during early stages of adversarial training. 
\\\\
\subsubsection{Gaussian Scales}
In similar spirit, we limit the range of predicted Gaussian scales:
\begin{align}
    \label{eq:scale_activation}
    \mathcal{G}_{scale}^i &\leftarrow \textsc{exp}\left(-s_{max} - \textsc{softplus}(-(G_{scale}^i - s_{init}) - s_{max})\right)
\end{align}
which ensures that scales are within $[0, e^{-s_{max}}]$ and initially default to $e^{-s_{init}}$ when the network predicts zeros. We set $s_{max} = 3$ and $s_{init} = 5$ which initializes Gaussian scales with $\sim$7mm roughly covering the template mesh, and limits them from exceeding $\sim$5cm.

\subsection{Gaussian Geometry Regularization}
\label{sec:asdf}
\label{sec:03_gaussian_geometry_reg}

Simply predicting 3D Gaussians in an adversarial setting leads to poor results. The reasons for this are two-fold:
(i) The 3D Gaussians react very sensitively to gradient updates during early training stages where the discriminator is not calibrated yet. 
(ii) The 3D Gaussian representation is quite powerful. Therefore, the generator is capable of predicting scenes that look plausible when rendered, but the underlying geometry lacks realism. \\
We find that a set of regularizations that implement simple intuitions effectively stabilize training as well as improve the underlying geometry. To tackle training stability, we use a simple L2 regularization on the predicted position offsets and Gaussian scales. To improve the geometric fidelity of generated 3D representations, we propose to use a beta regularization on the predicted offsets as well as a novel total variation loss on UV renderings.

\subsubsection{Gaussian Position Regularization} 

A common cause of instabilities during training are Gaussians moving around too much. We prevent this behavior with a position regularization on predicted offsets:
\begin{align}
    \mathcal{L}_{reg}^{pos} = \Vert M_{position} \Vert_2 
\end{align}
This term encourages predicted Gaussians to stay close to the template mesh.

\subsubsection{Gaussian Scale Regularization}

As opposed to implicit scene representations, such as NeRFs, 3D Gaussians are very sensitive to just a few parameters. For example, it is possible for a single Gaussian to grow tremendously in scale, covering the whole view frustum with just one primitive. Such cases are detrimental for training stability. We employ a simple regularization on the predicted Gaussian scales to avoid such degenerate 3D representations:
\begin{align}
    \mathcal{L}_{reg}^{scale} = \Vert M_{scale} \Vert_2
\end{align}
In combination with~\cref{eq:scale_activation} this regularization pushes Gaussian scales to be close to $\sim$7mm which is just large enough that Gaussians would cover the template mesh. 

\subsubsection{Gaussian Opacity Regularization}

To improve the geometric fidelity of the generated Gaussian representation, we employ a beta regularization on the opacities, encouraging that Gaussians should be either fully transparent or fully opaque:
\begin{align}
    \mathcal{L}_{reg}^{opac} = \textsc{Beta}(M_{opacity})
\end{align}
where $\textsc{Beta}$ is the negative log-likelihood of a Beta(0.5, 0.5) distribution~\cite{lombardi2019neuralvolumes}. 

\begin{figure*}[tb]
    \centering
    \includegraphics[width=\linewidth]{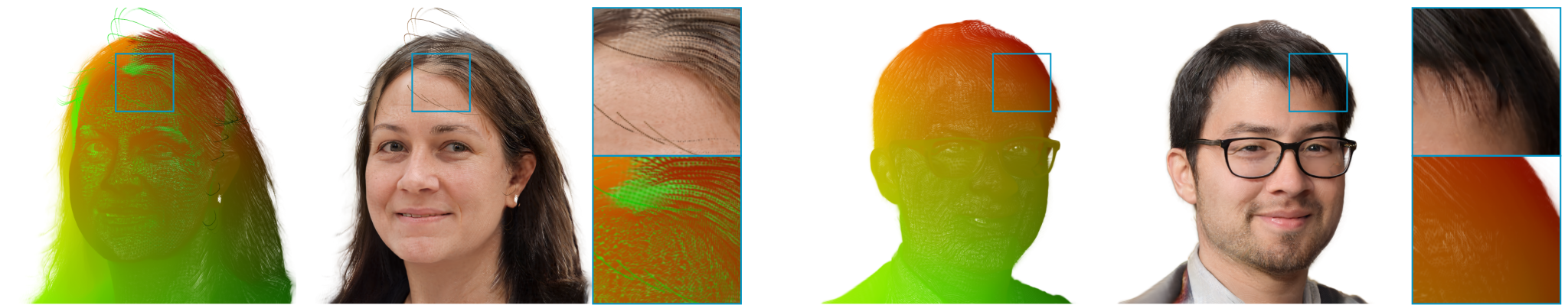}
    \begin{tabularx}{\linewidth}{CCP{0.08\linewidth}P{0.01\linewidth}CCP{0.08\linewidth}}
        \small$R_{uv}$ & \small{Rendered Image} &
        && \small$R_{uv}$ & \small{Rendered Image} & \\[3pt]
        \multicolumn{3}{c}{without $\mathcal{L}_{uv}$} 
        && \multicolumn{3}{c}{Ours}
    \end{tabularx}
    \caption{\textbf{Effect of TV UV regularization:} Without $\mathcal{L}_{uv}$, the predicted Gaussian geometry is flawed. Two common failure cases are the emergence of floating lines of Gaussians that are especially visible in video renderings, and improper surfaces where skin texture is created by letting Gaussians from the back of the head shine through. Both cases are easily detectable in the UV renderings $R_{uv}$. Our novel UV total variation loss exploits this fact and effectively addresses both issues by enforcing $R_{uv}$ to be smooth.}
    \label{fig:03_uv_tv_reg_effect}
\end{figure*}

\subsubsection{UV Total Variation Loss}

A naive implementation of our method generates Gaussian representations with many degrees of freedom. As a result, it can happen during training that the generator ``fakes'' high-frequency detail by letting colors from further back shine through. This still generates high-quality images but causes noticeable artifacts when rendering videos. As the discriminator is not aware of the time dimension, this behavior is never punished during training. We therefore design a novel regularization scheme to improve the geometric fidelity of generated representations. The intuition behind our regularization is that neighboring pixels in a rendered image should be modelled by Gaussians that are also close in UV space. This naturally leads to smooth surfaces and penalizes holes. Formally, we apply a total variation (TV) regularization on rendered UV images as follows:
\begin{align}
    \mathcal{\hat{G}} &\leftarrow \mathcal{G} \\
    \mathcal{\hat{G}}_{color}^i &\leftarrow (u^i, v^i, 0) \\
    R_{uv} &= \mathcal{R}(\mathcal{\hat{G}}, \pi) \\
    R_{uv}' &= \frac{R_{uv} - (1 - R_\alpha)}{R_\alpha} \label{eq:alpha_unblending} \\
    \mathcal{L}_{uv} &= TV(R_{uv}')
\end{align}
where $u^i,v^i = x^i_{uv}$ are a Gaussian's UV-coordinates, $\pi$ is the camera, and $R_\alpha \in \mathbb{R}^{H\times W}$ is the rendered alpha map containing the per-pixel accumulations from rasterization. We refer to $R_{uv}$ as a \textit{UV rendering}. \Cref{eq:alpha_unblending} reverses the alpha compositing that is performed during rasterization and is necessary in order to disregard transparency for the regularization. Otherwise, the border region between foreground and background, which is quite likely to include semi-transparent pixels due to the fuzzy nature of Gaussians, would contribute consistently to the regularization term, incentivizing foreground Gaussians to cover the background. With our novel UV total variation loss in place, holes in the 3D representation are stitched during training, noticeably improving the generated geometry. As a positive side effect, the regularization also removes other types of artifacts such as floating Gaussians that can appear in front of the face. \Cref{fig:03_uv_tv_reg_effect} shows how the generated 3D representation is improved by our novel regularization term. 

\begin{figure*}[tb]
    \centering
    \setlength{\tabcolsep}{0pt}
    \begin{tabularx}{\linewidth}{P{10pt}X}
         \rotatebox[origin=l]{90}{\parbox[c]{2.7cm}{\centering AFHQ-M} \parbox[c]{5.4cm}{\centering FFHQ-M}} & \includegraphics[width=\linewidth]{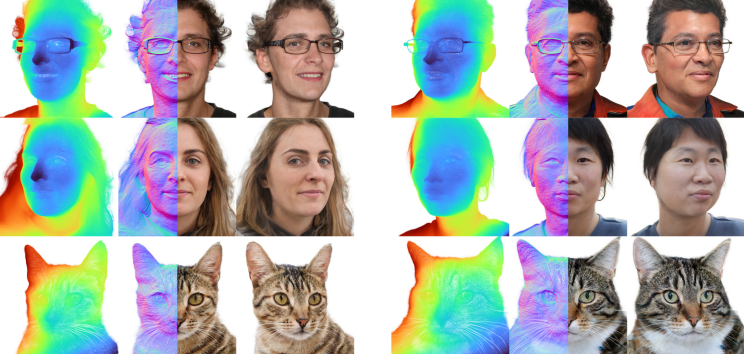} \\
    \end{tabularx}
    
    \begin{tabularx}{\linewidth}{P{15pt}YYYP{0.065\linewidth}YYY}
        & depth & \mbox{\,\,normals | rendering} & rendering &
        & depth & \mbox{normals | rendering} & rendering 
    \end{tabularx}
    \caption{\textbf{Generated Geometry:} We show samples from our models trained on FFHQ-M and AFHQ-M on $512^2$ resolution with corresponding depth and normal maps. {\OURS} can generate fine geometric details such as eyeglasses and hair strands.}
    \label{fig:04_geometry_results}
\end{figure*}

\subsection{Training Pipeline}
We generate UV maps $M$ at $256^2$ resolution and initially sample one Gaussian per UV map texel resulting in $65k$ Gaussians. We begin training at $256 \times 256$ rendering resolution and apply progressive growing after $7000k$ training images. Since the generator is resolution agnostic, we only need additional layers for the discriminator when increasing the rendering resolution. For progressive growing, we add a CNN block in front of the discriminator with a skip connection to ensure a smooth transition and blend in the contributions from the untrained layers over the course of $1000k$ images. We also increase the number of Gaussians when increasing the rendering resolution to facilitate modelling more high-frequency details. To achieve this, we simply increase the sampling density of Gaussians in UV space from $256^2$ to $512^2$ for a total of $262k$ Gaussians. This sudden increase in Gaussians causes the predictions to become temporarily more blurry. Nevertheless, training stability is not adversely affected by that and the generator quickly adapts to the presence of more Gaussians. Also note that the Gaussians are never all active at the same time for any given input latent $z$. The generator uses the opacity attribute to switch Gaussians on and off as needed, avoiding topological issues that would be implied by the template mesh. In practice, around half of the Gaussians are active on average. \\
The final optimization term for the generator is as follows:
\begin{align}
    \mathcal{L}_G &= \mathcal{L}_{adv} + \lambda_{p} \mathcal{L}_{reg}^{pos} + \lambda_{s} \mathcal{L}_{reg}^{scale} + \lambda_{o}\mathcal{L}_{reg}^{opac} + \lambda_{uv}\mathcal{L}_{uv} \\
    \mathcal{L}_{adv} &= \textsc{softplus}(-D(\mathcal{R}(\mathcal{G}), \pi))
\end{align}
where $\mathcal{L}_{adv}$ is the standard non-saturating GAN loss~\cite{goodfellow2014gan} applied on the rendered 3D Gaussian representation. The discriminator $D$ is trained with R1 gradient regularization~\cite{mescheder2018r1gan} with weight $1$. For the regularization weights, we choose $\lambda_p = 0.1$ and $\lambda_s = 0.05$. The opacity and UV total variation loss are initially turned off for training at $256^2$ resolution and are set to $\lambda_o = 1$ and $\lambda_{uv} = 100$ as soon as progressive growing is applied.
We use a batch size of $32$ with a learning rate of $0.0025$ for the generator and $0.002$ for the discriminator discriminator using the Adam optimizer~\cite{kingma2014adam}. We also incorporate the speed improvements by~\citet{durvasula2023distwar} for faster differentiable rasterization of Gaussians. In total, we train our model for 25M images following~\citet{chan2022eg3d} which takes roughly 12 days on 2 RTX A6000 GPUs with 48G of VRAM each.  \\ 

\section{Experimental Results}

\begin{figure*}
    \centering
    \includegraphics[width=\linewidth]{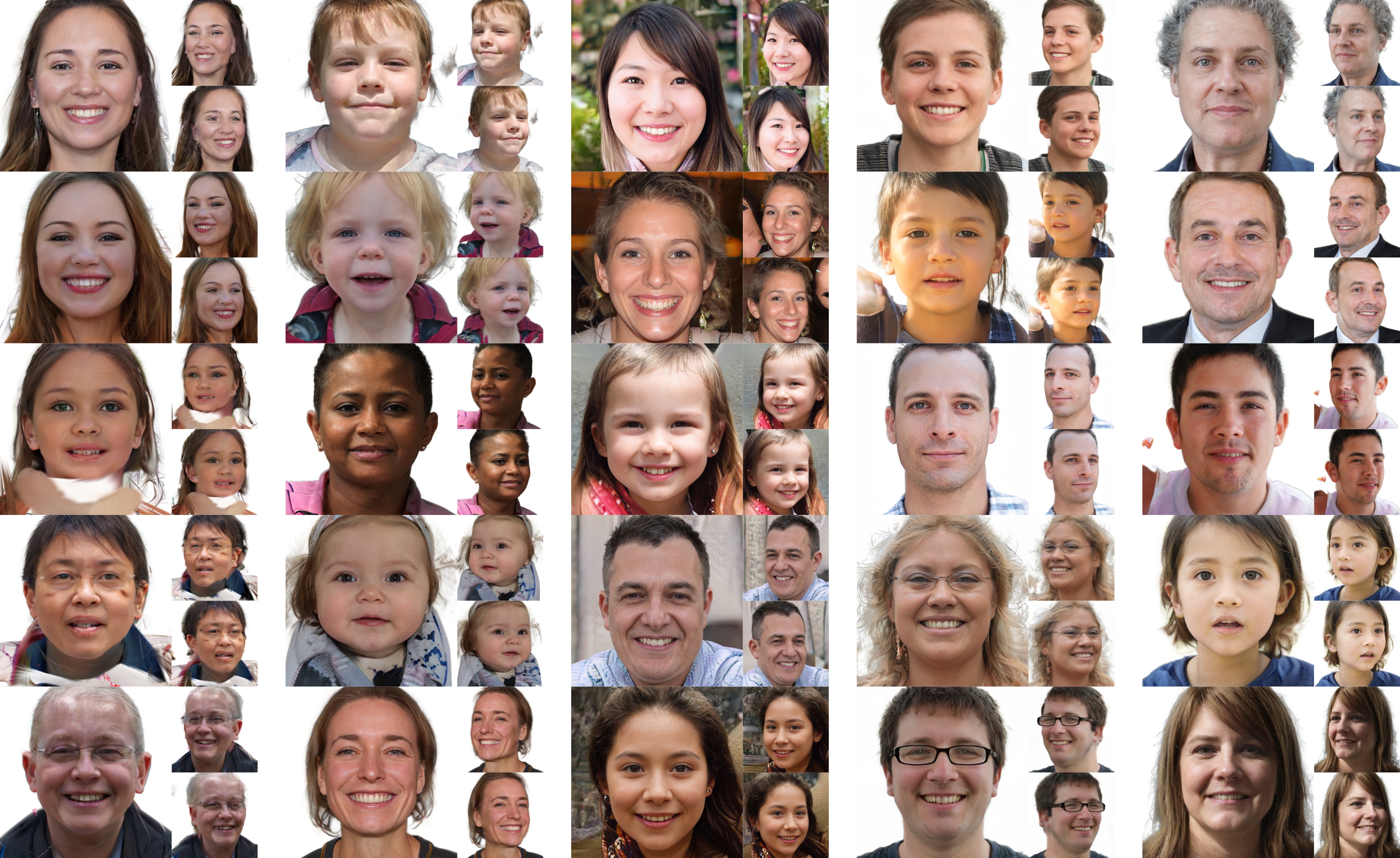}
    \begin{tabularx}{\linewidth}{YYYYY}
        \small{188 fps} & \small{42 fps} & \small{25 fps} & \small{3.5 fps} & \small{228 fps} \\
        Gaussian Shell Maps & EG3D & GRAM-HD & Mimic3D & Ours \\
    
    \end{tabularx}
    \caption{\textbf{Qualitative Comparison:} We show uncurated samples (seeds 0-4) of different 3D GANs trained on FFHQ at $512^2$ resolution. Our method matches the quality of existing approaches while being strictly 3D-consistent and one order of magnitude faster to render. Times measured on an RTX A6000 GPU with a batch size of 1. \\\\}
    \label{fig:04_results_comparison}
\end{figure*}

\subsection{Datasets}
We train our models on data from the FFHQ dataset~\cite{karras2019ffhq}, which consists of 70k mostly frontal images of human faces. We use the preprocessed version of~\citet{chan2022eg3d} who aligned the images via facial landmarks, cropped them into $512^2$ resolution, and computed camera poses by resorting to a 3DMM fitting~\cite{deng2019deep3dfacerecon}. We further follow EG3D's processing pipeline to obtain a $1024^2$ resolution dataset of cropped and aligned face images with corresponding camera poses from the original FFHQ in-the-wild dataset. Finally, we employ the off-the-shelf background matting network MODNet~\cite{ke2022modnet} to replace the background with white pixels yielding a masked version of FFHQ denoted as FFHQ-M. That way, our models can focus their generation power solely on the actual 3D head. We also apply the same masking strategy to the AFHQv2 Cats dataset~\cite{choi2020afhq} yielding AFHQ-M.

\begin{figure*}[tb]
    \centering
    \includegraphics[width=\linewidth]{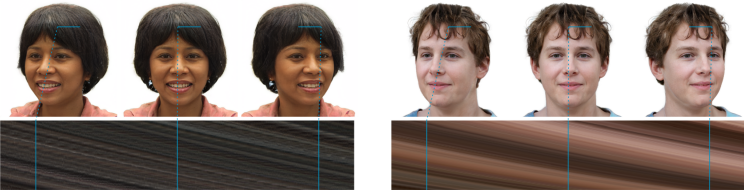}
    \begin{tabularx}{\linewidth}{CC}
        EG3D~\cite{chan2022eg3d} & Ours
    \end{tabularx}
    \caption{\textbf{Analysis of 3D Consistency:} We show spatio-temporal line textures akin to the Epipolar Line Images (EPI)~\cite{bolles1987epipolar} obtained by rotating the camera horizontally and stacking the pixels of a fixed horizontal line segment. 
    EG3D suffers from texture-sticking artifacts due to its use of a 2D super-resolution network, leading to staircases in the epiploar line images. In contrast, our method provides smooth renderings without any flickering.}
    \label{fig:04_temporal_consistency}
\end{figure*}

\begin{figure*}
    \centering
    \includegraphics[width=\linewidth]{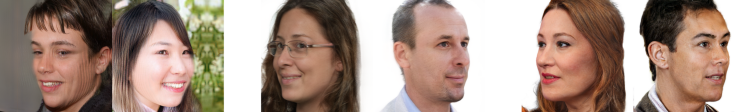}
    \begin{tabularx}{\linewidth}{YP{0.01\linewidth}YP{0.01\linewidth}Y}
        GRAM-HD~\cite{xiang2023gramhd} && Mimic3D~\cite{chen2023mimic3d} && Ours
    \end{tabularx}
    \caption{\textbf{Analysis of Side-views:} We show renderings of GRAM-HD, Mimic3D and our model from extreme side-views. GRAM-HD exposes the peculiarities of its underlying 3D representation in this scenario since the predicted 2D manifolds often run in parallel, causing striped artifacts. Mimic3D suffers from blurred colors when viewed from the side. In contrast, the 3D Gaussian representation generated by our method still looks appealing. }
    \label{fig:04_comparison_side_views}
\end{figure*}

\subsection{Baselines}
We compare our method to several competitive baseline methods from the 3D GAN literature. The baselines are selected based on three criteria: quality, speed/resolution, and architectural similarity.
\\

\textit{EG3D~\cite{chan2022eg3d}} is a state-of-the-art 3D GAN that synthesizes high-quality face images by relying on a super-resolution network to upsample low-resolution renderings from generated TriPlanes.
\\

\textit{Mimic3D~\cite{chen2023mimic3d}} is SoTA among fully 3D-consistent GANs on FFHQ. It distills a TriPlane generator backbone by directly supervising rendered images with pseudo-GT multi-view images obtained from EG3D.
\\

\textit{GRAM-HD~\cite{xiang2023gramhd}} proposed an efficient radiance field representation based on 2D manifolds, enabling faster rendering and training at $1024^2$ resolution.
\\

\textit{Gaussian Shell Maps~(GSM)~\cite{abdal2023gsm}} shares important architectural design decisions with our method, most notably that it also predicts 3D Gaussians in the UV space of a template.
\\\\
For EG3D and Mimic3D, we use the official implementation and retrain their methods on the masked FFHQ and AFHQ datasets. Since Gaussian Shell Maps was originally proposed for animatable full body synthesis, we adapt their implementation and retrain it on the FFHQ-M dataset. For GRAM-HD, we use official model checkpoints to conduct qualitative and timing comparisons.

\begin{table}[tb]
    \centering
    \caption{\textbf{Quantitative comparison:} We compare our method to other 3D GANs on FFHQ and AFHQ at $512^2$ resolution. FFHQ-M and AFHQ-M denote the resepective datasets with the background removed. Our method fares favorably against other fully 3D-consistent methods and comes close to the performance of EG3D despite not using any 2D super-resolution. {\OURS} also produces more consistent 3D surfaces while simultaneously being considerably faster.}
    \setlength{\tabcolsep}{3.5pt}
    \begin{tabular}{llrrrrr}
        \toprule
            &
                & \multicolumn{3}{c}{\small{FFHQ-M}} 
                & \small{FFHQ} 
                & \small{AFHQ-M}\\
            \cmidrule(lr){3-5} \cmidrule(lr){6-6} \cmidrule(lr){7-7}

                & Method 
                & \small{FID$\downarrow$} 
                & \small{$\mathrm{PSNR}_{mv}$$\uparrow$}
                & \small{$\mathrm{SSIM}_{mv}$$\uparrow$}
                & \small{FID$\downarrow$} 
                & \small{FID$\downarrow$} \\
            
        \midrule
            \multirow{1}{*}{\centering \small w/ SR} 
                & EG3D %
                    & \textbf{3.28}
                    & 28.74
                    & 0.899
                    & \textbf{4.70}
                    & \textbf{2.82}
                    \\
        \cmidrule(lr){1-7}
           \multirow{3}{*}{\centering \small w/o SR}
                & GSM %
                    & 28.19
                    & 26.40
                    & 0.919
                    & -
                    & - 
                    \\
                & Mimic3D 
                    & 4.27
                    & \underline{30.95}
                    & \underline{0.949}
                    & 5.37
                    & 4.48 %
                    \\
                & Ours 
                    & \underline{4.06}  %
                    & \textbf{33.29}
                    & \textbf{0.964}
                    & \underline{5.15}  %
                    & \underline{3.40}  %
                    \\
                
        \bottomrule
    \end{tabular}
    \label{tab:fid_table}
\end{table}

\subsection{Quantitative Comparison}

We conduct quantitative comparisons with EG3D, Mimic3D, and Gaussian Shell Maps on the FFHQ and AFHQv2 Cats datasets at $512^2$ resolution. For comparisons on the unmasked FFHQ dataset, we extend our generator to predict 3 additional channels to produce a background image and simply blend that with the rendered Gaussian head. For AFHQ-M, we take a model trained on FFHQ-M and finetune it on the much smaller cats dataset following~\citet{chan2022eg3d}. Since there is no direct ground truth to evaluate the quality of generated 3D representations, we resort to measuring the Fréchet Inception Distance~(FID)~\cite{heusel2017fid}. To do that, we generate 50k samples for each trained model using the training camera distribution and compare its statistics to the full dataset. The results are shown in~\cref{tab:fid_table}. Our method matches the performance of current state-of-the-art 3D GANs, obtaining slightly better FID scores than Mimic3D. It is only surpassed by EG3D, which employs a super-resolution module to achieve its rendering quality. To quantify the adverse effects of screen-space super-resolution on 3D consistency, we follow the procedure laid out by GRAM-HD~\cite{xiang2023gramhd}. For each method, we generate 30 images from different viewpoints of the same person, and then train the surface reconstruction method NeuS2~\cite{wang2023neus2} on the generated images. We then measure the reconstruction error with PSNR and SSIM scores. The reasoning behind this is that view-inconsistencies in the images given to NeuS2 will lead to a higher reconstruction error. In practice, we observe that this metric also measures ``surfaceness'' of a 3D representation, since NeuS2 favors smooth and well-defined surfaces by design. We report the multi-view surface reconstruction results averaged over 100 randomly generated heads in~\cref{tab:fid_table}. {\OURS} leads to significantly better surface reconstructions than EG3D, underscoring its great 3D consistency.

\subsection{Qualitative Comparison}
\Cref{fig:04_geometry_results} shows samples from our model alongside extracted depth and normal maps for both FFHQ-M and AFHQ-M. The generated 3D representations exhibit a great amount of geometric detail and diversity. We further compare samples from GGHead to those of EG3D, Mimic3D, GRAM-HD and GSM in~\cref{fig:04_results_comparison}. Our method produces renderings of great quality matching the current state-of-the-art at much higher speed. Interestingly, GSM synthesizes images with noticeable artifacts, indicating that generating 3D Gaussians alone is not sufficient for high-quality renderings. We attribute this to the design of GSM which disallows Gaussians from moving, severely restricting the expressiveness of the generated 3D representation. \\
Next, we analyse the view-consistency of generated images from EG3D and our method. EG3D generally can synthesize high quality static images, but this comes at the cost of view-inconsistencies due to the use of a 2D super-resolution network. In~\cref{fig:04_temporal_consistency}, we present the spatio-temporal texture images similar to Epipolar Line Images (EPI)~\cite{bolles1987epipolar} for EG3D and our method. Specifically, we smoothly rotate the camera around the generated head while stacking the pixels from a fixed horizontal line segment for each timestep. For multi-view consistent images, the resulting spatio-temporal texture should appear smooth. Conversely, edged artifacts and noise in the spatio-temporal texture indicate 3D inconsistencies. For EG3D, such artifacts are prevalent and manifest as flickering in video rendering. In contrast, our method generates strictly 3D-consistent images. \\
Furthermore, we investigate the generations of Mimic3D and GRAM-HD in more detail in~\cref{fig:04_comparison_side_views}. We find that both methods generally produce high-quality renderings but expose idiosyncracies of the underlying 3D representation when viewed from extreme angles. While GRAM-HD's use of 2D manifolds becomes visible in this scenario, Mimic3D produces slightly fuzzy surfaces, which is typical for NeRFs, causing blurry renderings for side views. Instead, our method produces more well-behaved renderings for such extreme camera angles. We attribute this to the fact that generating 3D Gaussian primitives in UV space forces them to lie on a single 2D manifold. This acts as an inductive bias towards more surface-like 3D representations which helps {\OURS} to generalize to such challenging scenarios. 

\subsection{Compute Resource Requirements}

\begin{figure}[tb]
    \centering
    \includegraphics[width=\linewidth]{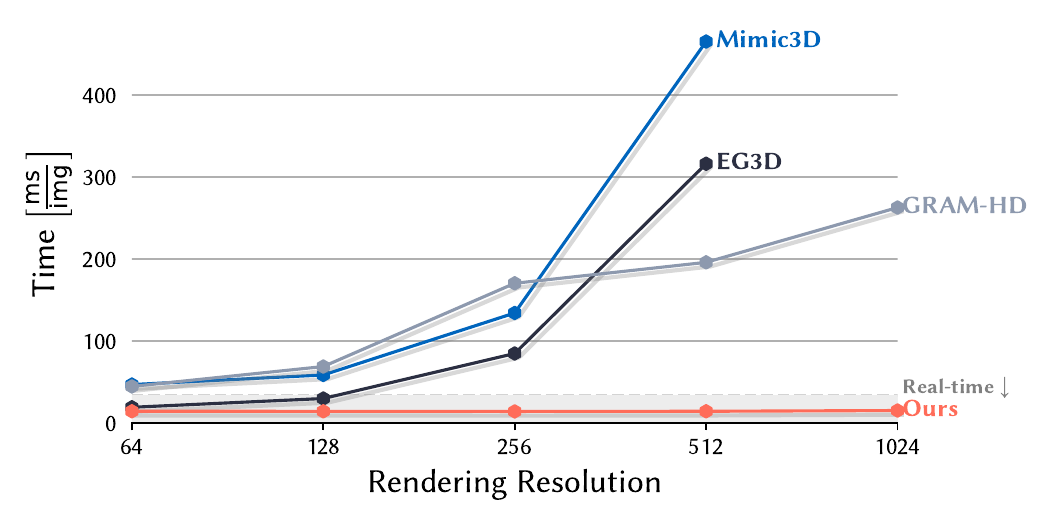}
    \caption{\textbf{Generation and Rendering Resolution Benchmark:} We measure the rendering times (including generation of the 3D representation) on a single RTX A6000 GPU with a batch size of 1. Our method can effortlessly generate high-quality 3D representations and render them at 1k native resolution in real-time while still matching other competitive 3D GANs in terms of FID.}
    \label{fig:04_benchmark_resolution_generation}
\end{figure}

A core motivation for our method is to lower the time needed for generating and rendering 3D heads, ultimately leading to faster training of 3D GANs at higher resolutions. We benchmark training and inference times of existing 3D GANs, and compare them to our method in~\cref{tab:04_benchmark}. As expected, we observe a huge performance gap between the NeRF-based approaches (EG3D, Mimic3D, GRAM-HD) and methods that employ 3D Gaussian Splatting (GSM, Ours). This is true not only for the actual rendering of the 3D representation itself, but also for the required time for a full training iteration where rendering also has to be performed. Note that due to expensive ray marching, EG3D and Mimic3D have considerably slower training iterations \textit{despite} rendering only $\frac{1}{16}$th of the pixels. \\
We further study the scalability of existing 3D GAN methods by measuring the required time for a full forward pass under different rendering resolutions. The scaling behavior is plotted in~\cref{fig:04_benchmark_resolution_generation}. Again, our method demonstrates a considerably better scaling behavior on larger rendering resolutions thanks to the efficient 3D Gaussian Splatting rasterizer. \\
Finally, we record GPU memory requirements and the total training time in~\cref{tab:04_memory_consumption}. Our method is significantly more memory-efficient than other 3D GANs and finishes training noticeably faster.

\begin{table}[tb]
    \centering
    \caption{\textbf{3D GAN performance benchmark:} We measure the time it takes to generate 3D representations (Gen.), render them (Rend.), and perform a full forward-backward pass during training (Forw. + Backw.). "Res." denotes the native rendering resolution \textit{excluding} super-resolution. Note that EG3D and Mimic3D have to lower the native rendering resolution substantially during training. In contrast, our method runs much faster at full resolution without sacrificing rendering quality. Times averaged over 1000 images.}
    \label{tab:04_benchmark}
    \begin{tabular}{lrrrrrr}
        \toprule
            & \multicolumn{2}{c}{Train} & \multicolumn{3}{c}{Inference} \\
            \cmidrule(lr){2-3} \cmidrule(lr){4-6}
            & {\small Res.} & {\small Forw. + Backw.$\downarrow$} & {\small Res.} & {\small Gen.$\downarrow$} & {\small Rend.$\downarrow$} \\
        \midrule
        EG3D
            & 128   & 209.61 ms     & 128   & \textbf{6.1 ms}   & 23.4 ms \\
        Mimic3D
            & 64    & 298.73 ms     & 512   & 20.3 ms  & 286.7 ms \\
        GRAM-HD
            & -     & -             & 512   & 157.6 ms & 38.2 ms \\
        GSM
            & 512   & \textbf{120.16 ms}    & 512   & 8.8 ms & 5.2 ms \\
        Ours 
            & 512   & 127.78 ms     & 512   & 10.4 ms & \textbf{3.8 ms} \\
        \bottomrule
    \end{tabular}
\end{table}

\begin{table}[tb]
    \centering
    \caption{\textbf{Compute Resource Requirements:} We measure the GPU memory consumption during training on a single RTX A6000 GPU (48GB VRAM). Out-of-memory is denoted as \textit{OOM}. Our method is significantly more memory-efficient and therefore scales better with larger batch sizes. We also record the total training time in equivalent days on a single RTX A6000.}
    \label{tab:04_memory_consumption}
    \setlength{\tabcolsep}{4pt}
    \begin{tabular}{lrrrrr}
        \toprule
            & \multicolumn{4}{c}{Batch Size} &
            \\

            \cmidrule(lr){2-5}
            Method 
                & 4 & 8 & 16 & 32 
                & Train time
            \\
            \midrule

            Mimic3D 
                & 20.4 GB & 38.7 GB & \textit{OOM} & \textit{OOM} & 80.2 days \\
            EG3D 
                & 13.4 GB & 25.1 GB & \textit{OOM} & \textit{OOM} & 39.0 days \\
            Ours 
                & \textbf{8.6 GB} & \textbf{12.8 GB} & \textbf{19.4 GB} & \textbf{35.0 GB} & \textbf{23.4 days} \\

        \bottomrule
    \end{tabular}
\end{table}

\begin{table}[tb]
    \centering
    \caption{\textbf{Ablation study:} We analyze the effect of regularization, different template meshes, different numbers of Gaussians, and our novel UV total variation loss when training our model on FFHQ-M.}
    \begin{tabular}{cclr}
        \toprule
         && Method & FID$\downarrow$ \\
        \midrule
        \multirow{7}{*}{\rotatebox[origin=r]{90}{\parbox[c]{1.5cm}{\centering 256}}}
            & \multirow{3}{*}{\rotatebox[origin=r]{90}{\parbox[c]{0.8cm}{\centering \small Regular-izations}}}
            & w/o $\mathcal{L}_{reg}^{scale}$ & collapse \\
            && w/o $\mathcal{L}_{reg}^{pos}$ & 10.92 \\ %
            && w/o zero-init layer $\mathcal{Z}$ & 6.71 \\ %
            \cmidrule(lr){2-4}
            & \multirow{4}{*}{\rotatebox[origin=r]{90}{\parbox[c]{0.8cm}{\centering \small Template}}}
            & Sphere Template & 4.59 \\
            && Plane Template & 4.27 \\
            && FLAME Template & 4.74 \\
            && Adjusted FLAME Template & \textbf{4.03} \\
        \midrule
        \multirow{3}{*}{\rotatebox[origin=r]{90}{\parbox[c]{0.2cm}{\centering 512}}}
            & \multirow{3}{*}{\rotatebox[origin=r]{90}{\parbox[c]{0.9cm}{\centering \small Gaussians}}}
            & 65k Gaussians & 6.39 \\
            && 262k Gaussians & 4.28 \\
            && + $\mathcal{L}_{uv}$ (Ours) & \textbf{4.06} \\

        \bottomrule
    \end{tabular}
    \label{tab:04_ablation_table}
\end{table}

\subsection{Ablations}

In the following, we analyze important design decisions of our method, including the choice of template mesh, the number of generated Gaussians, and the effect of our novel UV total variation loss.

\subsubsection{Effect of Regularization}
In the upper part of~\cref{tab:04_ablation_table}, we analyze the impact of the scale and position regularization terms $\mathcal{L}_{reg}^{scale}$ and $\mathcal{L}_{reg}^{pos}$, as well as the usefulness of the zero-initialized CNN layer $\mathcal{Z}$ that enforces small position offsets predictions in the beginning. We find that all measures improve training stability which translates to a better FID score. Without scale regularization, predicted Gaussians even disappear entirely leading to total training collapse.

\subsubsection{Effect of Template Mesh} 
\label{sec:04_effect_of_template_mesh}
In~\cref{tab:04_ablation_table}, we compare multiple versions of {\OURS} trained with different template meshes on the masked FFHQ dataset at $256^2$ resolution. We consider different types of template meshes, ranging from simple domain-agnostic shapes, like planes and spheres, to meshes that are designed for the face domain. Here, we compare two different head meshes: FLAME~\cite{FLAME:SiggraphAsia2017} and an adapted version of the FLAME mesh taken from~\citet{abdal2023gsm} which has a more efficient UV layout and the back of the head removed. It turns out that {\OURS} is fairly robust to the exact choice of template mesh since all templates produce reasonable results in terms of FID, with the adjusted FLAME mesh scoring best. We hypothesize that the efficiency of the UV layout and the density of allocated points in the frontal face region are more important than the actual shape of the template mesh. \\

\subsubsection{Effect of UV Total Variation Loss}

Being fundamentally a 3D point representation, the 3D Gaussians lack the notion of a proper surface. Consequently, it can happen during training that Gaussians in the face area become too small, allowing other Gaussians from further back to shine through. This can create high frequency detail that satisfies the discriminator loss and looks plausible in static renderings. However, such a 3D representation remains flawed and is easily exposed in video renderings. Since we cannot supervise the time dimension during training, however, we instead employ our novel UV total variation loss. The effect can be seen in~\cref{fig:03_uv_tv_reg_effect}. The regularization term effectively reduces the amount of Gaussians that shine through, considerably improving the surface quality of the generated 3D representations. As a side effect, the loss also resolves another failure case where lines of Gaussians would appear that mimic the appearance of hair strands but occur out of place. In~\cref{tab:04_ablation_table}, we also study the effect of the UV total variation loss quantitatively and find that it slightly improves FID.

\subsubsection{Effect of Number of Gaussians}

Due to our generator design, we can flexibly sample different numbers of Gaussians from the same model. We can use this to our advantage when progressively growing to larger resolutions where more high-frequency detail needs to be synthesized. As opposed to directly training with all Gaussians from the beginning, this has two advantages: Training is both faster and more stable due to fewer degrees of freedom. We analyze the effect of more Gaussians on the final results in~\cref{tab:04_ablation_table} where increasing the number of Gaussians (262k) gives a noticeable improvement in quality over keeping the same number of Gaussians (65k) as in lower resolution training.

\section{Limitations and Future Work}

We have demonstrated that {\OURS} can generate and render 3D heads at high quality. However, the generated 3D representations only provide a user with viewpoint control. Having additional control over the facial expressions would enable further use-cases such as 3D-consistent expression editing of single images. Since our method already employs a coarse template mesh, it stands to reason to extend it in similar fashion as \citet{qian2023gaussianavatars, sun2023next3d, abdal2023gsm}, e.g., by employing FLAME and training on FFHQ with corresponding expression codes. One step further, the improved scalability of our method may enable training on large facial video datasets~\cite{zhu2022celebv}, ultimately paving the way for building a photorealistic 3DMM from 2D data alone. \\
Another axis of improvement is generality. We have shown how a 3D generative model can be obtained with our pipeline for the narrow domain of human heads. However, extending our method to other domains such as ImageNet categories~\cite{deng2009imagenet} could enable learning more generic high-quality 3D priors. This could be done by having one template mesh per category and making the mesh itself learnable. As a result, the model could uncover suitable templates on its own. Additionally, an approach like that of~\citet{Skorokhodov2023imagenet3dgeneration} could be employed to drop the requirement for domain-specific keypoint detectors and alignment procedure.

\section{Conclusion}

In this work, we have proposed Generative Gaussian Heads (GGHead), a fast method for training 3D generative models for human heads on large 2D data collections. At the core of our approach lies the combination of a 3D GAN pipeline with efficient rendering from 3D Gaussian Splatting. We achieve this by parameterizing 3D Gaussian Heads as a set of 2D maps that live in the UV space of a template mesh and that can be efficiently predicted with powerful 2D CNN architectures. We further propose a novel UV total variation loss that exploits the design of our generator to improve geometric fidelity of generated 3D heads. In experiments on FFHQ, we demonstrate that our method matches the image synthesis quality of the current state of the art while being both fully 3D-consistent and considerably faster. In timing benchmarks, we confirm the great scalability of our method giving rise to real-time generation and rendering of photo-realistic 3D heads at $1024^2$ resolution.

\subsection*{Acknowledgements}
This work was supported by the ERC Starting Grant Scan2CAD (804724) and the German Research Foundation (DFG) Research Unit ``Learning and Simulation in Visual Computing''.
We would also like to thank Karla Weighart for proofreading and Angela Dai for the video voice-over.

\bibliographystyle{ACM-Reference-Format}
\bibliography{main}

\appendix

\end{document}